\documentclass[conference]{IEEEtran}
\IEEEoverridecommandlockouts
\usepackage{graphicx}
\usepackage{amsmath}
\usepackage{amssymb}
\usepackage{cite}

\makeatletter
\renewcommand{\@fnsymbol}[1]{\ifcase#1\or 1\or 2\or 3\or 4\fi}
\makeatother

\begin{document}

\title{GSMT: Graph Fusion and Spatiotemporal Task Correction for Multi-Bus Trajectory Prediction}


\author{%
Fan Ding$^{1,*}$, Hwa Hui Tew$^{1,*}$, Junn Yong Loo$^{1,\dagger}$, 
Susilawati$^{2,\dagger}$,\\ LiTong Liu$^{1}$, Fang Yu Leong$^{1}$, Xuewen Luo$^{1}$, Kar Keong Chin$^{3}$, Jia Jun Gan$^{3}$%
\thanks{$^{*}$These authors contributed equally to this work.}%
\thanks{$^{\dagger}$Corresponding authors (Emails: susilawati@monash.edu;
\protect\\ \hspace*{1em} loo.junnyong@monash.edu).}
\thanks{$^{1}$School of Information Technology, Monash University Malaysia.}%
\thanks{$^{2}$School of Engineering, Monash University Malaysia.}%
\thanks{$^{3}$Perunding Atur Trafik Sdn Bhd, Malaysia, which funded this research and whose support has been instrumental.}%
}

\maketitle

\begin{abstract}
Accurate trajectory prediction for buses is crucial in intelligent transportation systems, particularly within urban environments.  In developing regions where access to multimodal data is limited, relying solely on onboard GPS data remains indispensable despite inherent challenges. To address this problem, we propose GSMT, a hybrid model that integrates a Graph Attention Network (GAT) with a sequence-to-sequence Recurrent Neural Network (RNN), and incorporates a task corrector capable of extracting complex behavioral patterns from large-scale trajectory data. The task corrector clusters historical trajectories to identify distinct motion patterns and fine-tunes the predictions generated by the GAT and RNN. Specifically, GSMT fuses dynamic bus information and static station information through embedded hybrid networks to perform trajectory prediction, and applies the task corrector for secondary refinement after the initial predictions are generated. This two-stage approach enables multi-node trajectory prediction among buses operating in dense urban traffic environments under complex conditions. Experiments conducted on a real-world dataset from Kuala Lumpur, Malaysia, demonstrate that our method significantly outperforms existing approaches, achieving superior performance in both short-term and long-term trajectory prediction tasks.

\end{abstract}

\begin{IEEEkeywords}
Trajectory Prediction, Deep Learning, Graph Attention Network, Time-Series Modeling, GPS Data 
\end{IEEEkeywords}

\section{Introduction}
With the rapid growth of traffic demand, road capacity is increasingly strained. As an environmentally friendly, low-carbon, and efficient mode of transportation, the bus system can effectively alleviate the pressure on energy consumption and urban traffic. Therefore, traffic forecasting plays a crucial role in intelligent transportation systems (ITS). By extracting valuable information from historical data, traffic forecasting provides accurate predictions of future trends for better management and informed planning decisions \cite{zhang2024real,shanthi2024optimizing,li2023sequence}.

Bus trajectory forecasting is essential for system operators, enabling proactive responses to issues such as overcrowding, capacity constraints, and schedule disruptions. It also facilitates downstream tasks like arrival time estimation, passenger flow management, and traffic prediction. With advancements in bus hardware and GPS technology, real-time spatial data can now be readily collected via onboard sensors. Compared to costly multimodal data sources, GPS data is more accessible and offers a favorable cost-performance ratio for predictive applications.
Current vehicle trajectory prediction methods largely depend on time-series algorithms, including both traditional statistical models and deep learning approaches. However, accurately modeling bus trajectories remains challenging. Conventional methods often represent routes as linear sequences of stations, which perform well under normal conditions but fail to adapt to disruptions caused by congestion or adverse weather. Additionally, these models typically neglect spatial interactions across the network, limiting their ability to handle the dynamic and complex nature of real-world traffic.

To address the limitations of traditional methods, recent trajectory prediction approaches often leverage multimodal data, such as visual, auditory, and sensor inputs to improve accuracy. However, this strategy introduces practical challenges, including high data acquisition costs and significant computational overhead for data fusion and synchronization. These complexities hinder large-scale deployment, particularly in dynamic, real-world environments. 
Unlike single-vehicle prediction, forecasting the trajectories of an entire bus fleet on a given route requires modeling not only local interactions between adjacent buses but also the broader influence of upstream and downstream vehicles. Moreover, the model must generate short-term trajectory forecasts for all buses to enable operators to take timely interventions and prevent service disruptions.
To our knowledge, no existing work treats the entire multi-route bus fleet as a unified trajectory planning system.  This approach captures both direct and indirect bus interactions, enabling accurate future trajectory prediction and effective identification of issues like bus bunching.

Overall, the main contributions of our research are as follows:
First, we propose a hybrid sequence-to-sequence model based on multiple graphs for trajectory prediction of multiple buses along fixed routes, which can also be extended to multiple routes. The method relies solely on GPS data, providing a high-accuracy and lightweight solution for bus trajectory prediction. Subsequently, we construct a prediction task framework using a real trajectory dataset collected via GPS systems from buses operating in the busy urban areas of Kuala Lumpur. Experimental results show that the proposed model achieves an accuracy of 88.12\% for short-term predictions and 66.12\% for long-term predictions, outperforming existing models. Our work is the first to treat the same buses on a single bus route as a bus fleet while predicting the trajectories of the entire fleet by explicitly considering vehicle interactions.

\section{Related work}


Bus trajectory prediction plays a crucial role in public transportation systems, as it helps optimize bus scheduling, reduce passenger waiting times, and improve overall operational efficiency. As urban traffic complexity increases, traditional statistics-based approaches, such as historical averages or regression models, gradually show their limitations, especially when dealing with dynamic and complex traffic environments\cite{Ip2021,mo2021graph,jin2023spatio}. In recent years, deep learning methods, especially models combining Graph Neural Networks (GNN) and Recurrent Neural Networks (RNN), have provided more efficient and precise solutions for trajectory prediction tasks\cite{zhang2020crowd,li2021spatial,reza2022multi}.

\subsection{Spatial and Temporal Relationships in Modeling}
Bus trajectory prediction requires simultaneous modeling of spatial and temporal relationships. Spatial relationships typically refer to interactions among buses, traffic flow, and road conditions, which affect travel time\cite{mendez2023long,ma2019bus}. For example, GNNs model traffic networks as graphs, capturing spatial dependencies between nodes and effectively representing complex urban traffic environments\cite{chien2002dynamic,wang2022traffic,zhao2017lstm}. Temporal relationships, on the other hand, emphasize dynamic patterns, such as historical trajectories, peak traffic hours, and holidays, all of which influence future arrival times\cite{he2019stcnn,yao2019revisiting}. By integrating both spatial and temporal relationships, researchers have achieved significant improvements in trajectory prediction accuracy\cite{yu2017spatio,laptev2017time}.

\subsection{Graph Neural Networks (GNNs) in Trajectory Prediction}
GNNs have become an essential tool in processing traffic data due to their ability to capture spatial and temporal dependencies. By modeling traffic networks as graphs, GNNs address spatial heterogeneity and interconnectivity in predictive tasks\cite{vlahogianni2014short,li2017diffusion,zhang2017deep,tew2025st}. For example, Graph Attention Networks (GATs) dynamically adjust the relevance of neighboring nodes during the aggregation process using self-attention mechanisms, enabling them to capture complex relationships in graph-structured data\cite{rong2023gbtte,yang2024predicting,tew2024kans}. Multi-layer attention networks, as a subclass of GATs, further enhance this capability by stacking self-attention layers to model intricate inter-node correlations\cite{karimzadeh2021mtl,vishnu2023improving}.

\subsection{Time-Series Prediction Methods}
Recurrent Neural Networks (RNNs), particularly Long Short-Term Memory (LSTM) networks, are widely used for time-series prediction tasks due to their ability to capture temporal dependencies through recurrent structures\cite{li2023sequence,li2021transportation}. LSTMs have demonstrated impressive results in applications such as speech modeling, machine translation, and sequence-to-sequence forecasting \cite{yang2022ais,li2021semisupervised}. In bus trajectory prediction, LSTM-based methods, including BiLSTM (Bidirectional LSTM), effectively model long-term dependencies, improving prediction accuracy in complex urban traffic patterns\cite{aydemir2023adapt,mo2022multi}. Furthermore, adaptive LSTM frameworks have been proposed to integrate real-time traffic data and historical statistics for dynamic and accurate arrival time predictions\cite{lin2021spatial,capobianco2021deep}.

\subsection{Multi-Agent and Multi-Node Prediction}
Multi-agent trajectory prediction focuses on modeling the dynamic interactions among multiple trajectories simultaneously, which is critical in complex traffic environments. In interactive driving scenarios, frameworks incorporating traffic states have improved multi-agent trajectory prediction accuracy\cite{zhang2020crowd,mendez2023long}. These methods are particularly effective for multi-bus trajectory forecasting, enabling collaborative predictions in urban traffic\cite{li2021spatial,yao2019revisiting}.

\subsection{Lightweight Prediction Using GPS Data}
In resource-constrained scenarios, studies show that relying solely on GPS data can achieve high-accuracy trajectory predictions. Combining GPS data with Geographic Information Systems (GIS) further enhances prediction accuracy by simplifying multimodal data applications\cite{laptev2017time}.

\begin{figure}[t]
    \centering
    \includegraphics[width=0.9\linewidth]{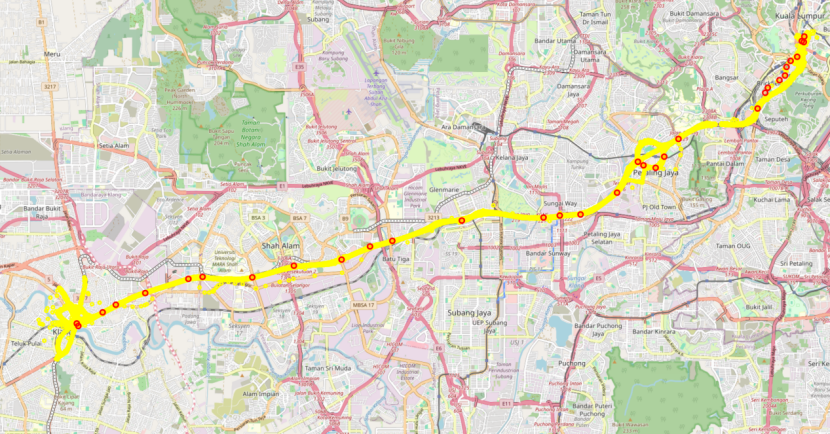}
    \caption{GPS trajectories of Route 710 in Kuala Lumpur from Prasarana Malaysia’s dataset (Jan–May 2024), capturing real-world latitude and longitude.}
    \label{fig:sql}
\end{figure}

\section{Preliminaries}
\subsection{GPS dataset}
The dataset, provided by Prasarana Malaysia, contains GPS data intermittently collected from onboard recorders on buses. It includes real-world latitude and longitude coordinates along with vehicle speed, covering operations from January to May 2024 without predefined schedules. For our experiments, we focus on Route 710, which runs through Kuala Lumpur’s dense urban traffic network, as illustrated in Fig.~1

\begin{figure*}[t]
\centering
\includegraphics[width=0.95\linewidth]{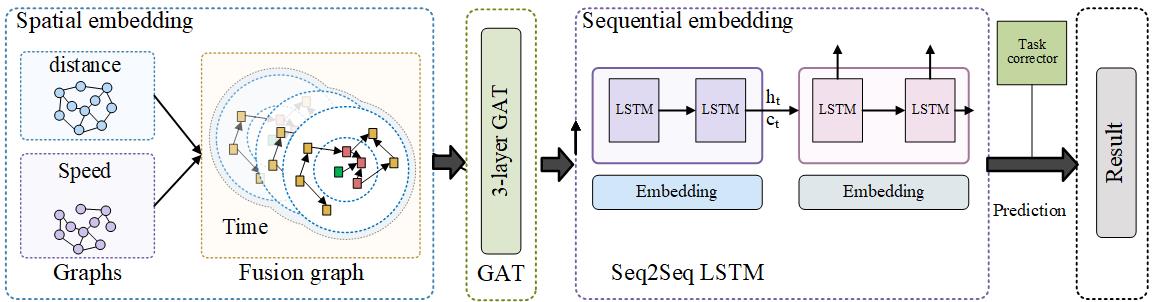}
\caption{The model integrates a 3-layer GAT, Seq2Seq LSTM, and a task correction module to achieve accurate trajectory prediction.}
\label{fig:sql2}
\end{figure*}

\subsection{Bus trajectory prediction}
Our input consists of a dynamic graph set formed by two identical nodes with different graphs. We define the feature vector of the node at time $t$ as $x(t) \in \mathbb{R}^{1 \times N}$, including coordinates of latitude and longitude as well as the velocity. We partition the entire set of dynamic graphs into a sequence of dynamic graphs, where each sequence \( S_i \) includes a fixed length of real values, which are used to feed into the model for training purposes.

\subsection{Task accuracy}
Task accuracy is calculated based on the task, and it can be deemed correct when the requirements are met. During the vehicle's journey, as the bus is in a state of high-speed travel and the time interval after data processing is one minute (i.e., the timestep is one minute), we calculate the average travel distance in the dataset. We take a 5\% error margin as the confidence interval; a trajectory falling within this interval is considered a correct prediction, while those falling outside this interval are considered incorrect.

\section{Proposed Methods}
In this section, we first provide an overview of the proposed model architecture, followed by a description on each of its components.

\subsection{Model architecture}
Our proposed model primarily consists of three components: the Graph Fusion Module, the Graph Attention Module, and the sequence-to-sequence RNN Module. As shown in Figure \ref{fig:sql2}, the Graph Fusion Module integrates features from multiple dynamic graphs and is highly scalable, allowing the fusion of historical information about roads and buses (nodes). The Graph Attention Module comprises a 3-layer Graph Attention Network (GAT), which dynamically learns the features of neighboring nodes and the topological relationships within the graph.  This generates richer node embeddings while encapsulating spatial structural information. 

A sequence-to-sequence RNN Module leverages an encoder to capture the contextual information of the input sequence and compress it into a hidden state, which is then used by the decoder to generate the output sequence and complete the final prediction. This architecture design allows the model to consider both the influence of nearby nodes and global nodes.
The use of LSTM as the sequence-to-sequence model, instead of traditional RNN or GRU architectures, enables the model to capture a greater amount of historical information.

\subsection{Dynamic graph fusion}

The first stage of our model focuses on fusing multiple dynamic graphs to comprehensively capture the spatial relationships between buses over time. As the model relies solely on GPS data, we directly aggregate information from multiple dynamic graphs, each representing the bus interactions at different time steps. Specifically, the adjacency matrices of multiple dynamic graphs are summed to generate a single fused adjacency matrix. This approach simplifies the prediction process by avoiding the need for excessive parameters and eliminates the complexity of processing each graph separately, thereby streamlining subsequent computations.

Formally, consider a sequence of $T$ dynamic graphs, where the adjacency matrix at time $t$ is denoted as $A^{(t)} \in \mathbb{R}^{N \times N}$, with $t = 1, 2, \dots, T$. The fused adjacency matrix $A^{\text{fused}}$ is computed as
\begin{equation}
    A^{\text{fused}} = \sum_{t=1}^T A^{(t)}.
\end{equation}
Each element of the fused adjacency matrix is given by
\begin{equation}
    a_{ij}^{\text{fused}} = \sum_{t=1}^T a_{ij}^{(t)}, \quad \forall i, j \in \{1, 2, \dots, N\},
\end{equation}
where $a_{ij}^{(t)}$ represents the weight of the edge between node $i$ and node $j$ in the graph at time $t$, and $a_{ij}^{\text{fused}}$ is the corresponding weight in the fused graph.

\subsection{Graph attention module}

After graph fusion, a unified graph is generated where buses are represented as nodes, and their relationships are represented as edges, incorporating all fused features. Each pair of buses operating on the same route is considered connected, with bidirectional edges.
To fully capture the spatial features in the input graph, we employ a three-layer Graph Attention Network (GAT) in the framework. The GAT extracts features from the dynamic graph sequence and propagates the processed graph features in sequence. During this process, GAT not only accounts for the influence of individual nodes but also considers interactions between nodes. Compared to traditional CNNs, GATs dynamically model the relationships between nodes and are well-suited for tasks involving dynamic or evolving graph structures.

For each pair of connected nodes \( l_i \) and \( l_j \) in the \( k \)-th layer, the correlation score \( e_{ij}^{(k)} \) is computed using a single-hidden-layer MLP as
\[
e_{ij}^{(k)} = \text{Attn}^{(k)}\left( \mathbf{u}_i^{(k-1)}, \mathbf{u}_j^{(k-1)}, \gamma_{ij} \right),
\]
where 
\( \mathbf{u}_i^{(k-1)} \) and \( \mathbf{u}_j^{(k-1)} \) are the features of nodes \( i \) and \( j \) from the \((k-1)\)-th layer, \( \gamma_{ij} \) represents the connection status between \( i \) and \( j \), and \( \text{Attn}^{(k)} \) is a single-hidden-layer MLP.
The correlation score \( e_{ij}^{(k)} \) is normalized into an attention weight \( \alpha_{ij}^{(k)} \) via a softmax layer as
\[
\alpha_{ij}^{(k)} = \frac{\exp(e_{ij}^{(k)})}{\sum_{k \in \mathcal{N}_i} \exp(e_{ik}^{(k)})},
\]
where \( \mathcal{N}_i \) is the set of neighbors of node \( l_i \).
The aggregated feature vector \( \mathbf{v}_i^{(k)} \) for node \( i \) is computed as a weighted sum over its neighbors via
\[
\mathbf{v}_i^{(k)} = \sum_{j \in \mathcal{N}_i} \alpha_{ij}^{(k)} \cdot \mathbf{u}_j^{(k-1)}.
\]
The node feature \( \mathbf{u}_i^{(k)} \) in the \( k \)-th layer is updated via a two-layer fully connected network with residual connection as
\[
\begin{aligned}
\mathbf{u}_i^{(k)} = \mathbf{u}_i^{(k-1)} + 
\sigma \big( W_2^{(k)} \cdot \sigma \big( W_1^{(k)} \cdot \mathbf{v}_i^{(k)} + \mathbf{b}_1^{(k)} \big) + \mathbf{b}_2^{(k)} \big),
\end{aligned}
\]
where 
\( W_1^{(k)}, W_2^{(k)} \) are the weight matrices of the fully connected layers, \( \mathbf{b}_1^{(k)}, \mathbf{b}_2^{(k)} \) are the biases, and $\sigma$ is the RELU activation function.
The GAT stacks multiple attention layers, and in the proposed model, three attention layers are used.

\subsection{Sequence-to-sequence LSTM module}\label{AA}
LSTM is widely regarded as significantly more effective than standard RNNs and GRUs in handling long-term dependencies. Unlike traditional LSTM models, our proposed approach employs a sequence-to-sequence LSTM architecture, which incorporates both historical and predicted information into each prediction output.
The encoder processes the input sequence step by step and compresses its information into a fixed-length context vector \( c \). At each time step \( t \), the encoder LSTM updates its hidden state \( h_t^{\text{enc}} \) and cell state \( C_t^{\text{enc}} \) as
\[
h_t^{\text{enc}}, C_t^{\text{enc}} = \text{LSTM}(x_t, h_{t-1}^{\text{enc}}, C_{t-1}^{\text{enc}})
\]
The final hidden state \( h_T^{\text{enc}} \) and cell state \( C_T^{\text{enc}} \) are passed to the decoder as the initial states.
The decoder generates the output sequence \( y \) step by step, conditioned on the encoder's context vector and its previous outputs. 

Similarly, the decoder LSTM updates its hidden state \( h_t^{\text{dec}} \) and cell state \( C_t^{\text{dec}} \) using the same LSTM update rule as the encoder.
The decoder's output at each step is computed as
\[
y_t = \text{softmax}(W_y h_t^{\text{dec}} + b_y)
\]
After each iteration of the prediction task, we obtain continuous predicted values for the vehicle trajectories.

\subsection{Task corrector}\label{AA2}

The task corrector is constructed based on historical data to identify complex motion patterns and refine the output sequence accordingly. It consists of a classifier that estimates the vehicle's current driving state and categorizes it into three representative motion modes: low-speed, medium-speed, and high-speed. For each category, the predicted trajectory is adjusted by modifying the trajectory computation. Specifically, the final output is refined by analyzing the direction aligned with the velocity vector, resulting in corrected predictions.

\begin{figure}[t]
\centering
\includegraphics[width=0.925\linewidth]{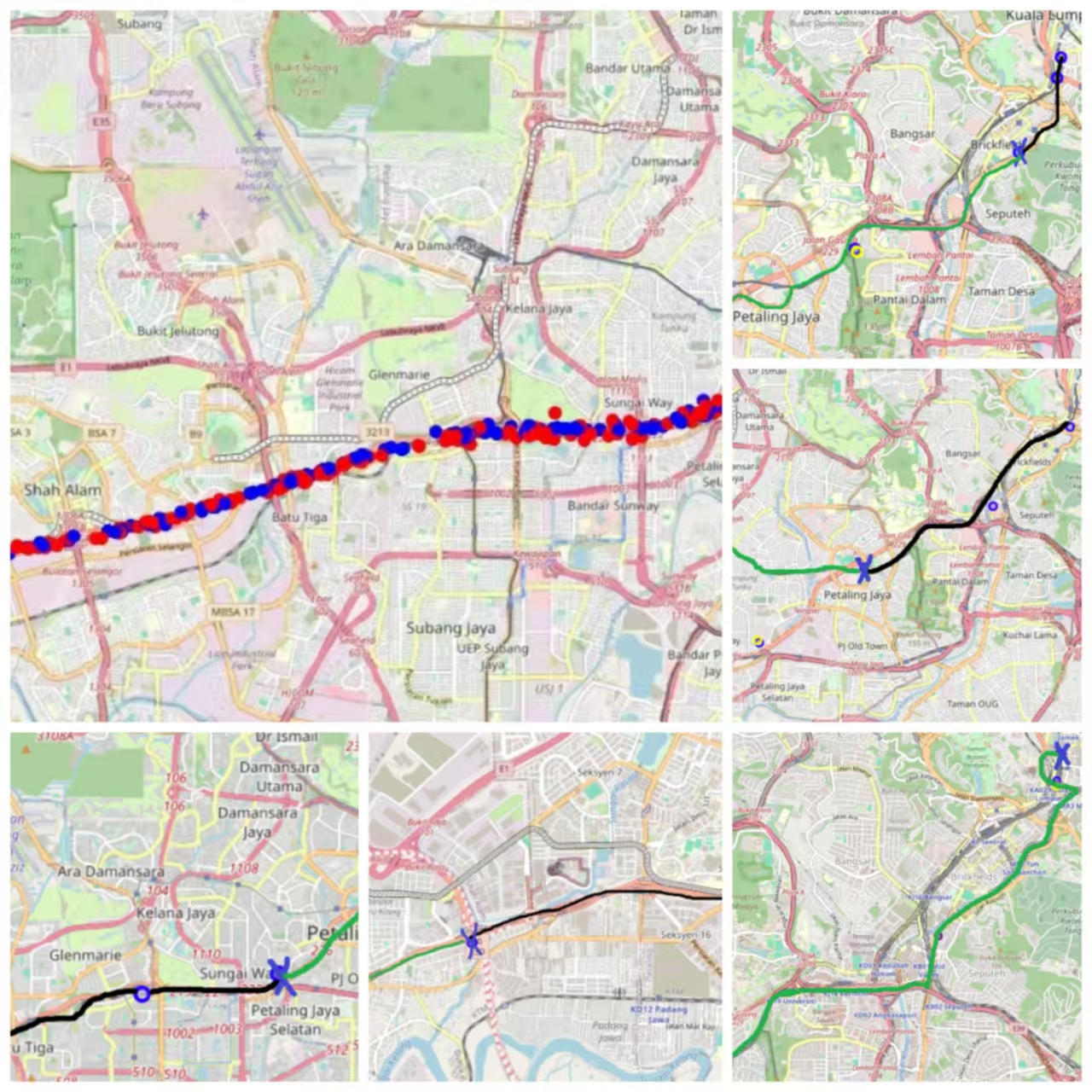}
\caption{Visualization of the bus routes in the test set and model prediction.}
\label{fig:single_column_image}
\end{figure}

\begin{table*}[t]
\centering
\caption{Comparison of forecasting models over two different forecast horizons (15 and 25 Minutes).}
\resizebox{0.75\linewidth}{!}{  
\begin{tabular}{lcccccc}
\hline
\textbf{Method} & \multicolumn{2}{c}{\textbf{15 Minutes  Forecast}} & \multicolumn{2}{c}{\textbf{25 Minutes Forecast}} \\ 
& \textbf{MAE} & \textbf{Mission Accuracy} & \textbf{MAE} & \textbf{Mission Accuracy} \\
\hline
Gat+HA & 0.6494 & 30.97\% & 0.4866 & 18.12\% \\
Gat+LSTM & 0.1427 & 46.12\% & 0.2236 & 35.16\% \\
Gat+GRU & 0.0605 & 77.92\% & 0.1660 & 63.19\% \\
Gat+transformer & 0.1406 & 47.42\% & 0.2453 & 23.36\% \\
Ours & 0.0515 & 88.12\% & 0.1510 & 66.12\% \\
\hline
\end{tabular}
}
\label{tab1}
\end{table*}

\section{Experiment}
In the experiments, we utilized real-world datasets to validate the effectiveness of the proposed architecture. Through data engineering, the complex dataset was transformed into a more learnable format. Additionally, we conducted comparisons with existing commonly used bus trajectory prediction methods on the same dataset. The results demonstrate that our approach achieves synchronized multi-agent predictions and outperforms others in terms of prediction accuracy.

\subsection{Data engineering}

As introduced in the dataset description, we extracted GPS data collected from one of Kuala Lumpur's busiest urban bus routes, utilizing five regularly operating buses as nodes. A review of the data revealed that these five buses represent the highest transport capacity on the route. This route exemplifies the complex and congested conditions of urban bus traffic. The dataset was collected with non-uniform sampling frequencies by different vehicles, and the lack of an exact timetable for the buses further increases the complexity of the data, making predictions more challenging.

Before feeding the data into the prediction model, we generated two distinct data graphs from the dataset, corresponding to different features: vehicle positional relationships and speed. During preprocessing, we removed sampling points beyond a certain range and extreme outliers, which were assumed to result from GPS sensor errors. Additionally, missing data points were imputed to minimize errors caused by missing values. To achieve finer-grained trajectory segmentation and meet task requirements, we set the time step to one minute. A 5-minute average step was used to replace the uneven sampling intervals, with the data at each time point represented by the corresponding average value.

Each time step is averaged and missing values are interpolated; the data is split into training (80\%), validation (10\%), and test (10\%) sets, with a sliding window used for input--output sequence segmentation.

\subsection{Experimental Setup}
In our experiments, we designed both short-term and long-term prediction tasks to validate the effectiveness of our model. For both tasks, a 5-minute time step was used, representing the coordinates of the bus at the current sampling moment. These coordinates correspond to real-world latitude and longitude values. To improve model accuracy, we first normalized the coordinates during data preparation, and the results were inverse-normalized after prediction to evaluate task precision.

The short-term prediction task uses 50 minutes of historical data to forecast the next 15 minutes, enabling the model to predict short-range bus trajectories. The long-term prediction task leverages 50 minutes of historical observations to forecast the subsequent 25 minutes, capturing longer-term trends in bus movements.

Each input consists of a sequence of 13 or 23 fused dynamic graphs, each containing 5 nodes, while the output comprises 3 or 5 dynamic graphs also with 5 nodes. The node features here represent the longitude and latitude information of the five buses.

During training, the loss was calculated using the Mean Absolute Error (MAE) between the predicted results and the actual results. The Adam optimizer was used for model optimization, and the training was conducted on a machine equipped with a single RTX 4090 GPU. We manually fine-tuned the hyperparameters during training to ensure the model achieved optimal performance during testing.

\subsection{Baseline}
At present, many studies related to bus prediction focus not on directly predicting the trajectories of multiple buses but rather on estimating bus arrival times. For baseline selection, we chose commonly used multi-graph methods for bus arrival time prediction and extended them to multi-node prediction using graph-based techniques. This adaptation ensures that the baseline methods achieve the same functionality as our proposed model for comparison.
Currently, the most commonly used bus-related prediction methods are graph-based and can be categorized into the following four types:
\textbf{HA}: Predicts vehicle trajectories based on the historical average data for each timestep\cite{fan2015dynamic}.
\textbf{GAT+LSTM}: Employs Graph Attention Networks combined with Long Short-Term Memory (LSTM) networks for sequence-to-sequence modeling\cite{alam2021predicting,wei2022convolutional}.
\textbf{GAT+GRU}: Utilizes Graph Attention Networks combined with Gated Recurrent Unit networks for sequence-to-sequence modeling.
\textbf{GAT+transformer}: Graph Attention Networks combined with transformer. 

\section{Result}

Through experimental validation, our proposed model demonstrates the highest accuracy and testing MAE in both short-term and long-term prediction tasks, as shown in Table \ref{tab1}. Additionally, we conducted validation in an offline environment with buses on a route. The proposed model is capable of simultaneously predicting the trajectory information of all buses operating in the bus network, and it performs well.
As shown in Table \ref{tab1}, we observe that traditional time-based methods exhibit almost no predictive ability for this task. Moreover, commonly used frameworks such as STGCN and T-GCN perform poorly when handling complex datasets. In contrast, our model maintains high accuracy in long-term prediction tasks with no significant performance degradation.

We also visualize the prediction results for a randomly selected batch from the test set in Figure \ref{fig:single_column_image}. In the main plot (top left), red indicates predicted values and blue shows the ground truth, demonstrating that the model successfully predicts complete trajectories for all nodes during this period. The remaining plots display individual time steps, where the black line shows the historical trajectory, the green line shows the predicted future trajectory, and the blue × marks the current vehicle position. For clarity, only the three most recent historical trajectories are retained in the visualization.

\section{Conclusion}
We have proposed a novel bus trajectory prediction model designed for complex scenarios, which leverages GPS sampling data to treat all buses along a given route as graph nodes for prediction. Unlike existing models, our approach combines a hybrid neural network architecture with historical data indicators, achieving high accuracy in predicting multi-vehicle trajectories in complex situations. 
Additionally, the architecture is inherently scalable and capable of expanding with graph networks to accommodate future demands.

\newpage

\bibliographystyle{IEEEtran} 
\bibliography{ref} 

\end{document}